\documentclass{article}
\usepackage{spconf,amsmath,graphicx,color}
\newcommand{\norm}[1]{\left\lVert#1\right\rVert_2}
\newcommand{\PG}[1]{\textcolor{black}{#1}}
\newcommand{\Isa}[1]{\textcolor{black}{#1}}

\newcommand{\GLB}[1]{\textcolor{black}{#1}}
\usepackage{enumitem}
\setlist{nosep, leftmargin=14pt}
\usepackage{mwe} 

\title{Automatic size and pose homogenization with spatial transformer network to improve and accelerate pediatric segmentation}
\name{\begin{tabular}{c}Giammarco La Barbera$^{1}$, Pietro Gori$^{1}$, Haithem Boussaid$^{2}$, Bruno Belucci$^{1}$ \\
Alessandro Delmonte$^{3}$, Jeanne Goulin$^{3}$, Sabine Sarnacki$^{3,4}$, Laurence Rouet$^{2}$ \\
Isabelle Bloch$^{1,5}$\end{tabular}}
\address{1- LTCI, Telecom Paris, Institut Polytechnique de Paris, France \and 2- Philips Research Paris, Suresnes, France \and 3- IMAG2, Imagine Institute, Universite de Paris, France \and 4- Universite de Paris, Pediatric Surgery Department, Necker Enfants-Malades Hospital, APHP, France \and 5- Sorbonne Universite, CNRS, LIP6, Paris, France}
\begin{document}
\maketitle
\begin{abstract}
Due to a high heterogeneity in pose and size and to a limited number of available data, segmentation of pediatric images is challenging for deep learning methods. In this work, we propose a new CNN architecture that is pose and scale invariant thanks to the use of Spatial Transformer Network (STN). Our architecture is composed of three sequential modules that are estimated together during training: (i) a regression module to estimate a similarity matrix to normalize the input image to a reference one; (ii) a differentiable module to find the region of interest to segment; (iii) a segmentation module, based on the popular UNet architecture, to delineate the object. Unlike the original UNet, which strives to learn a complex mapping, including pose and scale variations, from a finite training dataset, our segmentation module learns a simpler mapping focusing on 
\PG{images with normalized pose and size.}
Furthermore, the use of an automatic bounding box detection through STN allows saving time and especially memory, while keeping similar performance. We test the proposed method in kidney and renal tumor segmentation on abdominal pediatric CT scanners. Results indicate that the estimated STN homogenization of size and pose accelerates the segmentation \GLB{(25h)}, compared to standard data-augmentation \GLB{(33h)}, while obtaining a similar quality for the kidney \GLB{(88.01\% of Dice score)} and improving the renal tumor delineation \GLB{(from 85.52\% to 87.12\%)}.
\end{abstract}
\section{Introduction}
\label{sec:intro}
\begin{figure}[htb]
\begin{minipage}[b]{1.0\linewidth}
  \centering
  \centerline{\includegraphics[scale=0.8]{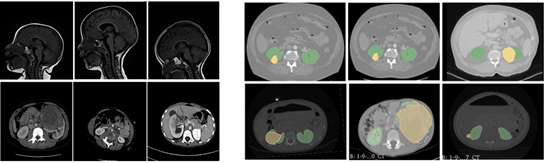}}
\end{minipage}
\caption{Left: Differences in size and pose in two of our pediatric data-sets. First row: MR sagittal brain images. Second row: CT axial abdominal images. Right: Differences between MICCAI KiTS19 \cite{kits19} adults images (first row) and our pediatric abdominal images (second row). Kidneys are in green and tumors in yellow.} 
\label{fig:diff-ped}
\end{figure}

Developing machine learning algorithms, and especially deep learning ones, for segmenting pediatric images is a challenging task. First, pediatric data-sets contain subjects going from few days of age to 16 years, showing therefore anatomical structures \textit{highly heterogeneous in terms of size}. Furthermore, due to the fact that children do not always stand still during the acquisition \cite{pediatric:images}, pediatric images also present a \textit{high variability in terms of pose} and movements artifacts.
Figure \ref{fig:diff-ped} (left) illustrates these problems.

Moreover, pediatric databases are limited in number of images~\cite{pedam} and therefore usual deep learning strategies might fail or not give good results \cite{big:data}. Direct inference or Transfer learning from 
networks trained on adults might fail for the differences between the two populations, especially in terms of relative size between organs and variability among subjects \cite{diff}. 
Some authors proposed to use an ad-hoc data-augmentation, as in \cite{ohbm}, to take into account the differences between adults and children. However, this usually takes time and it is not always possible or easy to recreate all the sources of variations (e.g. relative size between organs and tumors) in the data-augmentation process. 

For all these reasons, we propose to take a different perspective with respect to the usual data-augmentation strategy. Instead \GLB{of} 
\textit{augmenting} the number of training images to cover the entire data distribution, we propose to \textit{reduce} the data variability through an homogenization in terms of size and pose. In order to do that, we first \textit{learn} an optimal similarity transformation to a clinically relevant reference subject. Then, to accelerate the segmentation, we also \textit{learn} to crop the region of interest (ROI) as a square patch which is used as input image for the final segmentation network instead than the original (bigger) image.
We propose a new architecture composed of three neural networks: a first Spatial Transfomer Network (STN) \cite{stn} that deals with homogenization of pose and size; a second STN that crops the homogenized image in the region of interest (ROI); and finally a segmentation network, built as a nnUNet \cite{nnunet}, in which the cropped homogenized image is given as input and the output is then restored to its original pose and size, and uncropped, using the inverse of the two transformation matrices previously computed. This original combination allows to deal with small and heterogenous datasets, which is the main contribution of this work.
In this work, we focus on kidney and renal tumor segmentation on abdominal CT scanners which show high differences in tumor development between adults and children (Figure \ref{fig:diff-ped} on the right). Section~\ref{sec:sota} summarizes the state of the art in the segmentation of these structures. In Section~\ref{sec:data}, we describe our pediatric dataset. In section~\ref{sec:meth} the proposed method is detailed, and then, in Section~\ref{sec:rad}, results on 2D images are discussed.

\section{State of the art}
\label{sec:sota}

According to \cite{dl:mi}, the 3D extensions of U-Net \cite{unet:1} are the most used Deep Learning architectures for the segmentation of medical images, providing the best results.
\GLB{However,} to achieve high performance with 3D CNN, large datasets are needed \cite{big:data}, and currently most of the pediatric datasets do not contain \GLB{enough images}.
To overcome this limitation, transfer learning techniques from adults to children have been proposed \cite{ohbm,tl:aug}, but they usually require an ad-hoc and time-consuming data augmentation to take into account the anatomical variations between children and adults. \PG{For these reasons, 2D networks are usually chosen for pediatric datasets with less than 100 subjects.}
While the literature is poor on the specific problem of 2D pediatric kidney and renal cancer segmentation, recent works on 2D adult images are worth to be mentioned~\cite{kits19,crossbar}. No-newUNet~\cite{nnunet}, a framework implementing both 2D and 3D U-Net~\cite{unet:1}, is the network that manages to obtain the best results, thanks to the use of an important data augmentation. 
When working with pediatric images, the high variability in size and pose makes the distribution of data more heterogeneous compared to adult datasets. This entails a higher number of possible transformations in the data augmentation and therefore a more important computational time.
In~\cite{equi}, the authors propose to augment the convolutional kernels (instead of training data) by transforming them with several rotations. This allows the network to learn feature maps associated with different rotated versions of the input image in a single pass. However, variations in size could not be taken into account.

With a different perspective, we propose to learn a single and specific similarity transformation per image instead of computing many during training.
Each image is thus normalized in pose and size with respect to a reference image (relevant for clinicians). This simplifies the task for the segmentation network and avoids the computation of many time consuming transformations during the data augmentation.



\section{DATABASE}
\label{sec:data}
We worked on a pediatric dataset of abdominal-visceral CT images from 80 patients, with early arterial contrast injection. All patients presented a renal tumor and images were acquired pre-operatively. The age ranges from 3 years old to 16, with an average of 2 years old. Slices have a low pixel size (0.35 - 0.95 mm) with a size of 512$\times$512 pixels. Reference segmentations were performed by manual annotation under the supervision of medical experts.
These exams were performed in the course of the normal care pathway of the patient and were studied retrospectively after anonymization.
In the experiments, we also use the open access database KiTS19 \cite{kits19}, which collects 210 adults with renal tumor. All images are labeled by clinicians and acquired with the same medical technique of acquisition as ours. Even pixel size is comparable (0.65 - 0.95 mm).

\section{METHODS}
\label{sec:meth}

\begin{figure*}[htb]
%
  \centerline{\includegraphics[width=0.8\textwidth]{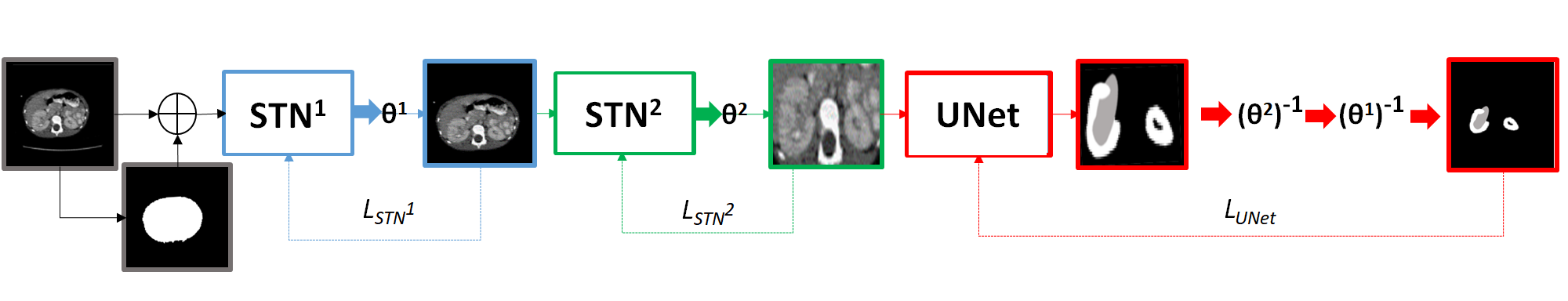}}
%
\caption{Schema of our proposed framework (see Section \ref{ssec:arc} for details).} 
\label{fig:framework}
\end{figure*}

\paragraph*{Pre-Processing}
\label{ssec:prepro}
All images are preprocessed using the tool ``pre-processing" of nnUNet \cite{nnunet}. The pre-processing consists of: (i) a non-zero region cropping, (ii) a resampling of the images to have the same pixel size, (iii) a clipping of the intensity values to the 0.5 and 99.5 percentile of the foreground voxels, and (iv) a Z-scoring normalization. 


\paragraph*{Architecture}
\label{ssec:arc}
The framework is presented in Figure \ref{fig:framework}. We now present the three networks in detail.

\paragraph*{STN to homogenize pose and size}
\label{sssec:stnh}
At first a Spatial Transfomer Network (STN) deals with homogenization, transforming all images to be as similar as possible in size and pose to a chosen one (STN\textsuperscript{1} in Figure \ref{fig:framework}). The reference image was chosen among patients aged 2 years, who represent the average in the database, and among them a patient with the best pose was chosen, according to the doctors' directives. This STN is composed of a localization network, composed of an encoder with two stacked convolutional blocks with MaxPooling and ReLU, which reduces the image by a factor of 4, and 2 fully convolutional layers. This regresses five values (1 value for angle, 2 for scaling and 2 for translation), defining the similarity matrix $\theta$\textsuperscript{1} which is then applied to a grid in which the starting image is interpolated thanks to a sampler. The input of the STN is composed of the original image concatenated with its ``foreground mask", a binary mask representing the abdomen and easily computed as the largest connected component.
The network is optimized using a Soft Dice loss function $L_{STN^1}$ between the homogenized output ``foreground mask" and the ``foreground mask" of the reference image:
\begin{equation}
\resizebox{.9\columnwidth}{!}{
$L_{STN^1} = SoftDice (\theta^1 I,T) = SoftDice (H,T) = \frac{2|H \cap T|}{|H|^2+|T|^2}\cdot 100 $
\label{eq:softdice}
}
\end{equation}
where $I$ is the input ``foreground mask", $\theta^1$ is the predicted matrix, $H$ is the homogenized output ``foreground mask" and $T$ is the reference ``foreground mask".

\paragraph*{STN for ROI cropping}
\label{sssec:stnc}
Then, there a second STN crops the homogenized image in the region of interest (ROI), where the structures to be segmented are present (STN\textsuperscript{2} in Figure \ref{fig:framework}). This network is the same as the previous one but it regresses 4 values, corresponding to the vertices of the bounding box, that are used to construct a scaling and translation matrix $\theta$\textsuperscript{2} for cropping.
A target matrix and the associated target bounding box are automatically calculated using the minimum and maximum non-zero values of the reference segmentation. However, the bounding box is forced to be a square and the minimum crop size is considered to be a quarter of the original image. This allows not deforming the image too much.
We underline that in our method the user can choose whether to keep the image in its original size, halve it or reduce it to a quarter (minimum size of the patches coming out of the STN). This allows, as mentioned in Section \ref{sec:intro}, reducing time and memory requested for the segmentation network.
The second STN for the cropping is trained using the loss function $L_{STN^2}$, defined as the sum of a $L^1$ loss (mean absolute error) between the cropped output image and the target crop, and a $L^2$ loss (mean squared error) between the ``scaling and translation” output matrix and the ``scaling and translation” target matrix:
\begin{equation}
\resizebox{.9\columnwidth}{!}{
$L_{STN^2} =\frac{1}{N} \sum_{n=1}^N \norm{{H_C}_n - {T_C}_n} +  \sqrt{\frac{1}{N} \sum_{n=1}^N \norm{\biggl(\theta^2_n - \theta^T_n\biggr)}^2}$
\label{eq:l1-l2}
}
\end{equation}
where $H$ is the homogenized input, $\theta^2$ is the predicted matrix, $H_C$ is the cropped output, $\theta^T$ is the target matrix, $T_C$ is the target crop and $N$ is the batch size. \Isa{This combination was proved experimentally efficient, probably due to the robustness of the $L^1$ norm to outliers.}

\paragraph*{U-Net for segmentation}
\label{sssec:unet}
At the end of our framework a U-Net takes as input for the segmentation the cropped homogenized image and the output is then restored to its original pose and size, and uncropped, using the inverse of the two transformation matrices previously calculated. The U-Net is constructed using the tool ``planes" of nnUNet, which suggests the best configuration and hyperparameters based on the input images. Each level is composed of two blocks, where the second differs between encoder and decoder part.
\begin{itemize}
    \item 1st block: 2D convolutional layer (kernel = 3, stride = 1, zero-padding = 1), batch normalization and ReLU; 
    \item 2nd block encoder: equal to first block except for stride = 2 to downsample the image by the same factor;
    \item 2nd block decoder: equal to the first block except for 2D transposed convolution layer instead of the classic convolution;
    \item output block: equal to the first block except for Softmax as activaction function.
\end{itemize}
The depth of the network is up to a bottleneck defined so to have a $8\times8$ image and skip connections are used up to the 32$\times$32 image level. The number of features is doubled with each downsampling in the encoder and halved during upsampling in the decoder. Our U-Net is optimized as in the nnUNet training using a loss function $L_{UNet}$ defined as the sum of cross entropy $CE$ and Soft Dice loss, both between
prediction and reference segmentation.

\begin{equation}
\resizebox{.9\columnwidth}{!}{
$L_{UNet} = \sum_{n=0}^N \frac{1}{2^n}\biggl(CE({\theta^1}^{-1}({\theta^2}^{-1}P_n),G) + SoftDice({\theta^1}^{-1}({\theta^2}^{-1}P_n),G)\biggr) $
\label{eq:unet}
}
\end{equation}
where $P$ is the prediction, $G$ is the reference segmentation and $n$ is the level of the prediction (considering the output layer of the network as 0 level). We use the Deep Supervision technique \cite{deepsup} up to the level $N$, where there is the last skip connection.


\paragraph*{Training}
\label{ssec:train}
\GLB{For STNs the best solution has been experimentally identified as a training of one STN after the other for 50 epochs using a Stochastic Gradient Descent (SGD) optimizer with a learning rate of 0.01}. 
The U-Net is trained for at least 50 epochs with a subsequent early-stopping condition, using a SGD with a starting learning rate of 0.01 with a poly learning rate policy \cite{plrp}, a Nesterov momentum of 0.99 and a weight decay of $3\times 10^{-5}$, \GLB{as nnUNet}. We used 15036 2D images for training and 3760 for validation, using an oversampling technique to have at least one third of the images containing the kidney and one third with tumor. 15 subjects for a total of 5310 slices are used as test set.

\section{RESULTS AND DISCUSSION}
\label{sec:rad}

In our first experiment, we used the 3D no-newUNet \cite{nnunet} trained on adults, winner of the KiTS19 challenge \cite{kits19}, directly on the children images (same weights), and then using transfer learning (fine tuning of the weights). The results, evaluated using the Dice Score in Table \ref{tab:adult-child}, show that only when we fine-tune most of the weights the results become satisfactory. This confirms the important differences between adults and children images, as shown in Figure~\ref{fig:diff-ped}.

\begin{table}[htbp]
  \caption{Results (mean and standard deviation of Dice score) using weights of 3D nnUNet  trained on adults KiTS database \cite{kits19}.}
  \label{tab:adult-child}
  \resizebox{\columnwidth}{!}{%
  \begin{tabular}{|l|c|c|c|c|c|c|c|c|}
    \hline
    Technique & Dice Score Kidney  & Dice Score Tumor \\ 
    \hline
    Direct Inference (weights frozen) & 20.83 (35.55)  & 18.29 (35.73) \\
    Fine-Tuning (first 2 blocks encoder and last 2 decoder) & 53.38 (25.84)
 & 51.05 (31.76) \\
    Fine-Tuning (entire decoder) & 81.75 (7.18) & 75.79 (23.24)\\
    Fine-Tuning (entire network) &  84.99 (6.38) & 81.08 (23.01)\\
    \hline
  \end{tabular}
  }
\end{table}

The next step was to test the size and pose homogenization $STN^1$ network on our pediatric database. We chose a 2D network since results based on a 3D network were not satisfactory due to the limited number of images available. At first, tests were carried out with the images resized to 128$\times$128. 
Then tests were made with the original size 512$\times$512. The results for both tests are shown in Table~\ref{tab:the_table}. The baseline is the original nnUNet, with and without the use of random data augmentation on-the-fly. The results show that the use of the $STN^1$ to homogenize pose and size outperforms \GLB{(increase of the mean and decrease of the standard deviation)} both the transfer learning 3D results (Table~\ref{tab:adult-child}) and the baseline with data augmentation for the tumor segmentation task, while showing comparable results for the kidney segmentation, especially on 512$\times$512 images. The slightly greater results in kidney segmentation are probably due to the use of a mirroring during data augmentation, not reproducible by our method.

\begin{table}[!htbp]
  \caption{Results (mean and standard deviation of Dice score and total traning time) on our pediatric database adding the proposed STNs to the baseline nnUNet (without data augmentation).}
  \label{tab:the_table}
  \resizebox{\columnwidth}{!}{%
  \begin{tabular}{|l|c|c|c|}
    \hline
     \multicolumn{4}{|c|}{\textbf{Image 128x128 with Batch Size of 32}}\\
    \hline
        Architecture & Training Time &  Dice Score Kidney  & Dice Score Tumor \\ 
    \hline
    nnUNet & 1h35 & 83.66 (7.88) & 69.52 (24.61) \\
    nnUNet (+ data augmentation) &  2h15 & \textbf{88.99 (3.71)} & 74.18 (22.07) \\
    STN pose-size + nnUNet & \textbf{1h45} & 86.75 (6.47) &	\textbf{77.31 (27.36)} \\
    \hline
         \multicolumn{4}{|c|}{\textbf{Image 512x512 with Batch Size of 12}}\\
   \hline
        Architecture & Training Time & Dice Score Kidney  & Dice Score Tumor \\
    \hline
    nnUNet & 22h & 88.07 (5.61) & 78.14 (26.19) \\
    nnUNet (+ data augmentation) & 33h & \textbf{88.91 (5.08)} & 85.52 (24.65) \\
    STN pose-size + nnUNet & \textbf{25h} &  88.01 (6.25) &	\textbf{87.12 (23.39)} \\
    \hline
  \end{tabular}
  }
\end{table}

\begin{figure}[htb]
\begin{minipage}[b]{1.0\linewidth}
  \centering
  \centerline{\includegraphics[width=8.5cm]{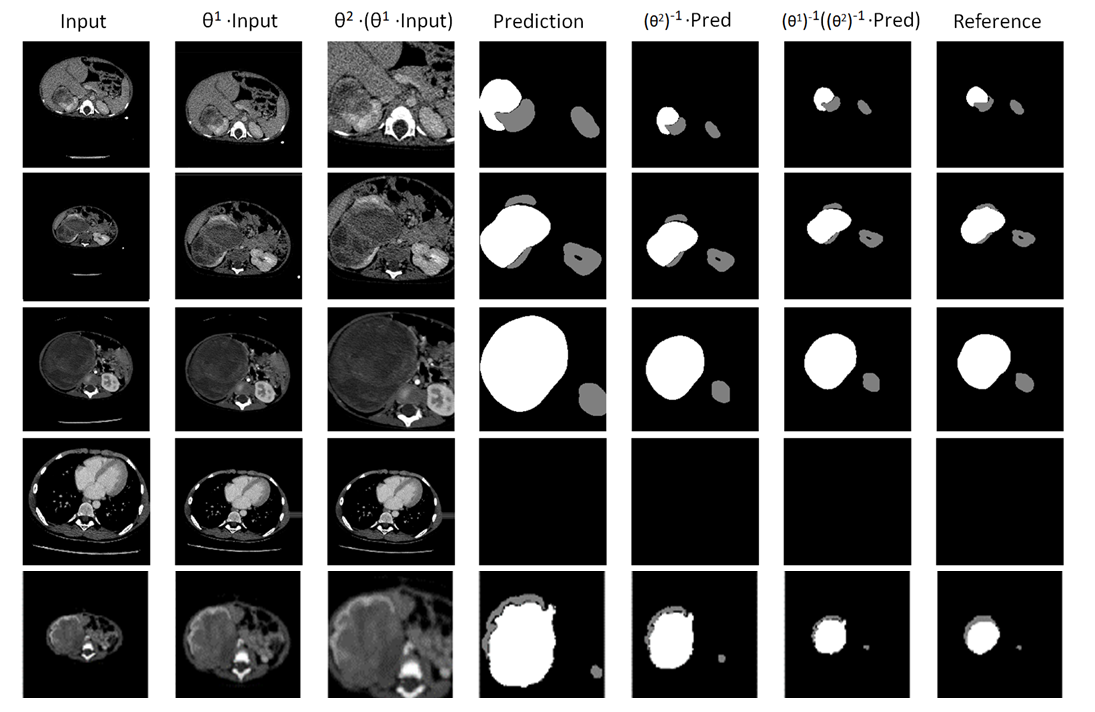}}
\end{minipage}
\caption{Qualitative results of our method illustrated step-by-step. All input images are 512$\times$512. In the last line, the cropped image is downsampled to 256$\times$256 and it can be noticed that the boundaries between tumor and renal cavities are lost.}
\label{fig:examples}
\end{figure}

In our case, the combination of the two $STNs$ does not lead to improvements in performance compared to using $STN^1$ alone, but it leads to a gain in time and requested memory as shown in Table~\ref{tab:table_time_memory} while maintaining high performance. This is due to the fact that the UNet has a smaller image as input.
The drop in performance depends on the renal tumor size, and consequently on the size of the ROI, which varies from $[128\times128]$ to $[380\times380]$. This means that, when reducing the input size of the UNet to $[256\times256]$ or $[128\times128]$, we actually downsample the ROI thus loosing important information, as shown in the last row of Figure~\ref{fig:examples}. Nevertheless, we believe that the proposed differentiable module to localize ROIs may be important for other datasets with smaller structures to segment compared to the size of the image (e.g. adults, see Figure~\ref{fig:diff-ped}), or when training time and memory are limited.


\begin{table}[!htbp]
  \caption{Results (mean and standard deviation of Dice score and total traning time) on our pediatric database reducing the size of the input image for nnUNet (memory allocated column refers only to nnUNet, STNs occupy less than 4Gb of RAM in the GPU also with 512$\times$512 inputs). Note that each network is trained individually}
  \label{tab:table_time_memory}
  \resizebox{\columnwidth}{!}{%
  \begin{tabular}{|l|c|c|c|c|c|}
  \hline
\multicolumn{1}{|c|}{Architecture} & \multicolumn{1}{c|}{\begin{tabular}[c]{@{}c@{}}Input size\\ UNet \end{tabular}} & \multicolumn{1}{c|}{\begin{tabular}[c]{@{}c@{}}Training\\ Time\end{tabular}} & \multicolumn{1}{c|}{\begin{tabular}[c]{@{}c@{}}Memory \\ allocated\end{tabular}} & \multicolumn{1}{c|}{\begin{tabular}[c]{@{}c@{}}Dice score\\ kidney\end{tabular}} & \multicolumn{1}{c|}{\begin{tabular}[c]{@{}c@{}}Dice score\\ tumor\end{tabular}} \\ \hline
    nnUNet & 512$\times$512 & 22h & 10.05Gb  & 88.07 (5.61) & 78.14 (26.19) \\
    nnUNet (+ \GLB{data aug.}) & 512$\times$512 &  33h & 10.05Gb   & 88.91 (5.08) & 85.52 (24.65) \\
    STN\GLB{p-s} + STNcrop + nnUNet & 512$\times$512 & 28h & 10.05Gb  & 88.84 (7.79) & 84.25 (31.15) \\
    STN\GLB{p-s} + STNcrop + nnUNet & 256$\times$256 & \textbf{19h30} & \textbf{3.52Gb}  &  86.71 (19.36) & 84.15 (30.11) \\
    \hline
  \end{tabular}
  }
\end{table}


In Figure~\ref{fig:examples}, results of the proposed network are illustrated step-by-step on images 512$\times$512. In the first four rows, we do not change the input size of the UNet, whereas in the last row we reduce it to 256$\times$256. This results in a less detailed image and thus a drop in performance.

\section{CONCLUSION}
\label{sec:conc}
In this work, we propose to use a Spatial Transformer Network as a method to reduce data variability on pediatric images through an homogenization of size and pose, improving performances and computational time with respect to standard data augmentation. Moreover, the use of a second STN to crop images around the structures to segment can save even more computational time and memory, while maintaining high performance. 
Future work aims to combine the two STNs into a single one and extend it to 3D (when a sufficient number of images will be available). 
\bibliographystyle{IEEEbib}
\bibliography{strings,refs}

\paragraph*{Compliance with Ethical Standards}
This research study was conducted retrospectively on images performed in the normal course of the care pathway of the patients. Ethical approval was not required.

\paragraph*{Acknowledgments}
This work has been partially funded by a grant from Region Ile de France (DIM RFSI).

\paragraph*{Conflict of interest}
The authors have no relevant financial or non-financial interests to disclose.
    
\end{document}